\documentclass[letterpaper, 10 pt, conference]{ieeeconf}  

\usepackage{amsthm}
\usepackage{graphicx}
\usepackage{float}
\usepackage{amsmath}
\usepackage{amssymb}
\usepackage{mathtools}
\usepackage{booktabs} 

\usepackage{flushend} 
\usepackage{cite}
\usepackage{wrapfig}
\usepackage{makecell}
\usepackage{tikz}
\usepackage{algorithm}
\usepackage{algorithmic}
\usepackage{tcolorbox}
\usepackage[font=footnotesize]{caption}
\usepackage{bbm}
\usepackage{subfig}
\usepackage{hyperref}
\usepackage{xcolor}

\hypersetup{
    colorlinks,
    linkcolor={blue!50!black},
    citecolor={blue!50!black},
    urlcolor={blue!80!black}
}

\theoremstyle{plain}
\newtheorem{theorem}{Theorem}[section]

\theoremstyle{definition}
\newtheorem{definition}[theorem]{Definition}

\theoremstyle{remark}

\IEEEoverridecommandlockouts                              

\title{\LARGE \bf
Dexterous Manipulation from Images: \\Autonomous Real-World RL via Substep Guidance
}

\author{Kelvin Xu$^{*1}$ Zheyuan Hu$^{*1}$ Ria Doshi$^{1}$ Aaron Rovinsky$^{1}$ Vikash Kumar$^{2}$ Abhishek Gupta$^{3}$ Sergey Levine$^{1}$
\\
$^{1}$ UC Berkeley $^{2}$ Meta AI Research $^{3}$ University of Washington
\thanks{*Both authors contributed equally}%
\thanks{\href{https://sites.google.com/view/dexterous-avail/}{https://sites.google.com/view/dexterous-avail/}}%
}

\DeclareMathOperator*{\argmax}{arg\,max}

\begin{document}

\newcommand{\methodname}{AVAIL}
\newcommand{\classifier}{p^o}
\newcommand{\taskgraph}{p(z|s)}
\newcommand{\E}{\mathbb{E}}

\makeatletter
\let\@oldmaketitle\@maketitle%
\renewcommand{\@maketitle}{\@oldmaketitle%
    \centering
    \includegraphics[width=.70\linewidth]{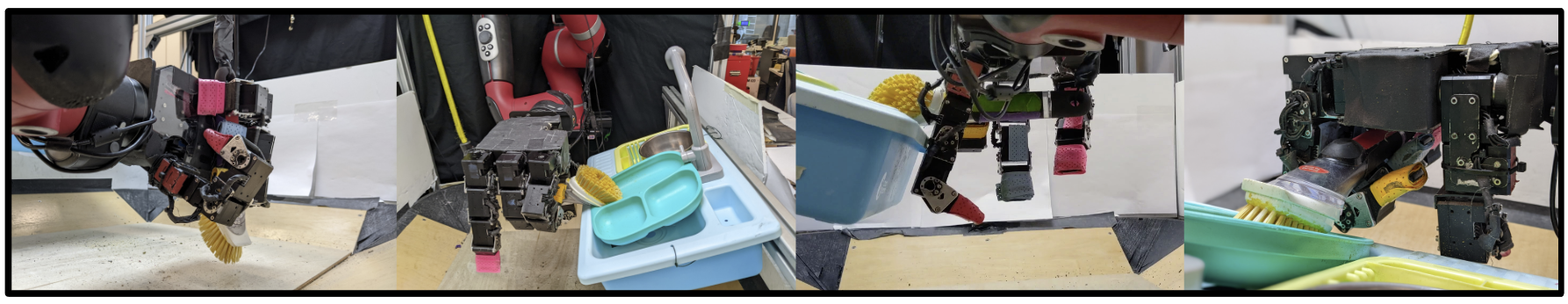}
    \vspace{-0.05in}
    \captionof{figure}{\footnotesize Filmstrip of the final learned brush skill. Our agent is able to learn to grasp, in-hand reorient, and brush a surface using a kitchen brush. After around $\sim 18$ hours of unattended learning, our system successfully performs each of these sub-tasks with $\geq 80$\% success.}
    \vspace{-0.15in}
    \label{fig:teaser}
}
\makeatother

\maketitle
\thispagestyle{empty}
\pagestyle{empty}

\begin{abstract}
Complex and contact-rich robotic manipulation tasks, particularly those that involve multi-fingered hands and underactuated object manipulation, present a significant challenge to any control method. Methods based on reinforcement learning offer an appealing choice for such settings, as they can enable robots to learn to delicately balance contact forces and dexterously reposition objects without strong modeling assumptions. However, running reinforcement learning on real-world dexterous manipulation systems often requires significant manual engineering. This negates the benefits of autonomous data collection and ease of use that reinforcement learning should in principle provide. In this paper, we describe a system for vision-based dexterous manipulation that provides a ``programming-free'' approach for users to define new tasks and enable robots with complex multi-fingered hands to learn to perform them through interaction. The core principle underlying our system is that, in a vision-based setting, users should be able to provide high-level intermediate supervision that circumvents challenges in teleoperation or kinesthetic teaching which allow a robot to not only learn a task efficiently but also to autonomously practice. Our system includes a framework for users to define a final task and intermediate sub-tasks with image examples, a reinforcement learning procedure that learns the task autonomously without interventions, and experimental results with a four-finger robotic hand learning multi-stage object manipulation tasks directly in the real world, without simulation, manual modeling, or reward engineering.
\end{abstract}

\section{Introduction}

Complex and contact-rich robotic manipulation tasks, particularly those that involve multi-fingered robotic hands and underactuated objects, present significant challenges to any control method. Reinforcement learning (RL) offers an appealing choice for such settings, as it in principle enables a robot to learn to adeptly apply contact forces and manipulate objects without strong modeling assumptions, directly from real-world experience. However, running RL on real-world robotic platforms raises a number of practical issues that lie outside the standard RL formulation, such as difficulties with reward specification, state estimation and the practicalities of autonomous training.
Addressing such issues typically requires significant manual engineering or human intervention. This has led researchers to study alternative solutions, such as transfer from simulation~\cite{tobin2017domain,peng2018sim,openAIhand}, imitation learning~\cite{argall2009survey,hussein2017imitation,osa2018algorithmic}, or use of cumbersome instrumentation, such as motion capture~\cite{peng2018sim,kumar2016learning}.
Even when these issues can be overcome, effective real-world reinforcement learning typically requires considerable reward engineering~\cite{mtrf}, complex reset mechanisms or scripts~\cite{pddm}, and other manually designed components.
Each of these solutions erodes the original benefits of autonomy and ease of use that RL should in principle provide. Solving even the simplest tasks with RL requires considerable domain and robotics expertise to program reward functions and reset mechanisms for autonomous operation. Thus, in order to allow for learning-based dexterous manipulation systems to reach their full potential in terms of practicality, accessibility, and scalability, it is critical to limit the assumptions on manual engineering while still providing enough supervision for reinforcement learning to be tractable.

In this work, we propose a robotic learning system that can learn to control high-dimensional multi-fingered robotic hands from raw visual observations, without the need for extensive engineering for every new task. In the absence of simulation, manual reward shaping, and hand-designed state estimation instrumentation, we aim to enable RL to be as autonomous as possible. The robot should be able to practice the task for a long period of time without human intervention, and the task itself should be specified in a way that does not require per-task programming or human-in-the-loop supervision. To this end, we propose a system that autonomously practices a sequence of sub-skills based on high-level milestone specifications provided by the user that break up a complex task into more manageable sub-problems. The milestone specifications consist of snapshots of critical states illustrated by posing the robot and objects in the scene. For multi-fingered hands, such examples are significantly easier to provide than full demonstrations, and our system can use them to learn reward functions that provide sufficient shaping for RL in the real-world without per-task engineering or specific reward design. The system uses multi-camera visual observations to localize and manipulate objects, with policies learned end-to-end from pixels and no motion capture. By sequencing the sub-skills appropriately and introducing very simple physical instrumentation (in our experiments, tethering the object to prevent it from falling out of reach), the robot can learn dexterous behaviors by practicing for up to 48 hours fully autonomously. The milestone decomposition makes both reward inference and autonomous practicing significantly easier, enabling real-world learning of complex tasks. Our experiments (see Fig.~\ref{fig:teaser} for an example) show that this approach can learn skills that involve basic grasping, in-hand reorientation, and object manipulation, through significant amounts of practicing, entirely from images and without task-specific reward engineering.

\section{Related Work}

Prior work has studied control of complex hands using trajectory optimization~\cite{mordatch2012contact,kumar2014real}, policy search~\cite{kober2008policy,posa2014direct, rajeswaran2017learning}, simulation to real-world transfer~\cite{openAIhand,lowrey2018reinforcement,allshire2021transferring}, and real-world reinforcement learning~\cite{van2015learning,zhu2019dexterous}. In contrast to our work, the majority of this prior work has assumed access to compact state representations or accurate simulators and object models. Closer to the system we describe in this paper is prior work on learning visuomotor policies for dexterous manipulation~\cite{jain2019learning,mandikal2020learning,akinola2020learning}
However, with the exception of some work we discuss below, prior systems on RL for dexterous manipulation typically require assumptions on manually designed rewards, or ground truth object state observations. These assumptions hinder the application of RL in more real-world settings.

An important consideration in our system is the ability to specify a task without manual reward engineering, by using intermediate milestone examples. Previously studied methods for task specification include having humans provide demonstrations for imitation learning~\cite{argall2009survey,imitation_learning_martial_hebert_drew_bagnell_uav,reddy2019sqil}, using inverse RL~\cite{ZiebartMBD08,wulfmeier2015maximum,maxmarginIRL}, active settings where users can provide corrections~\cite{losey2018including,cui2018active, coreyes2018lgpl}, or ranking-based preferences~\cite{myers2022learning,brown2020better}. While some prior work ~\cite{nachum2018hiro, levy2017hac} also uses subgoals, these are firstly restricted to reaching only particular goal states rather than more abstract milestones and are only applied in much simpler simulated problems with perfect state estimation. Motivated by the goal of broader applicability, we do not assume access to expert demonstrations (e.g., via teleoperation or kinesthetic teaching), which can themselves be difficult to provide for high-dimensional systems~\cite{akgun2012trajectories, villani2018survey}. For example, providing kinesthetic demonstrations for a full hand-arm robotic system requires very challenging coordination and several simultaneous demonstrators, and is incompatible with vision-based policy learning (as the demonstrator is in the scene, and often occludes the robot or objects). In contrast, we utilize sparse images of intermediate outcomes that can be obtained simply by positioning the robot and object in particular states and build on the VICE framework~\cite{fu2018variational} for reward inference. Our focus is not on devising a new \emph{algorithm} for learning rewards, but on leveraging existing components, such as VICE, to build a complete, autonomous, robotic system that can enable scalable RL with a dexterous manipulator in the real world. Therefore, although some of the building blocks of our system are based on prior work, their combination and the capabilities they enable (learning image-based dexterous manipulation in the real world) are novel.

The most closely related robotic RL systems that have been previously proposed are R3L~\cite{r3l} and MTRF~\cite{gupta2021reset}. Our assumptions regarding lightweight instrumentation and vision-based autonomous learning most resemble those of Zhu et al. (2020)~\cite{r3l} (R3L). However, our work tackles a considerably more challenging setting: while Zhu et al. (2020)~\cite{r3l} studied a 3-finger claw mounted on a fixed base, we show that our method can control a 4-fingered hand on a 7 DoF arm. This is done by leveraging a multi-task RL formulation that builds on ideas from MTRF~\cite{gupta2021reset} instead of the novelty-based resets in R3L~\cite{RND}, which scale poorly in higher dimensional settings. In contrast to these prior works, our focus is on providing a framework that can enable vision-based learning of object manipulation skills with high-dimensional hands via a lightweight milestone-based task specification mechanism. Part of this requires an automated RL system that can run continuously for 48 hours, though unlike MTRF~\cite{gupta2021reset}, we still employ lightweight physical instrumentation (by tethering the object to prevent it from falling outside of grasping range). We instead focus on the separate challenges of visual perception and reward specification, avoiding the need for manual reward engineering of the sort used by MTRF and completely circumventing the requirement for motion capture that was crucial in MTRF. We summarize these key system-level differences in Table~\ref{table:differences}.

\def\yes{\tikz\fill[scale=0.4](0,.35) -- (.25,0) -- (1,.7) -- (.25,.15) -- cycle;} 
\def\no{$\mathbf{\times}$}
\begin{table}[t]
\vspace{0.1in}
\caption{ A comparison between the assumptions of \methodname{} and prior autonomous RL methods.}\label{table:differences}
\centering
\resizebox{0.90\columnwidth}{!}{%
\begin{tabular}{ccccc}\toprule
\textbf{Method} & \makecell{\textbf{No Hand} \\ \textbf{Engineered Reward}} & \textbf{Multi-Task} & \textbf{Vision} & \textbf{High-DoFs}\\\midrule
R3L~\cite{RND} & \yes & \no & \yes & \no\\  \midrule
MTRF~\cite{gupta2021reset} & \no & \yes & \no & \yes\\  \midrule
Ours & \yes & \yes & \yes & \yes\\  \bottomrule
\end{tabular}%
}
\vspace{-0.3cm}
\end{table} 

\section{Robotic Platform and Problem Overview}
\label{sec:prelim}

We first present an overview of our robot platform, describing the hardware and task setup, as well as the observation and action space. Then, we provide an overview of our problem setting, focusing on the practical goals of our system. We provide complete details related to our robotic platform in our project website\footnote{\footnotesize{\href{https://sites.google.com/view/dexterous-avail/}{https://sites.google.com/view/dexterous-avail/}}} along with details of a simulated analogue that we employ for analysis and ablation experiments.

\begin{figure}
  \centering
  \vspace{0.1in}
  \includegraphics[width=0.99\columnwidth]{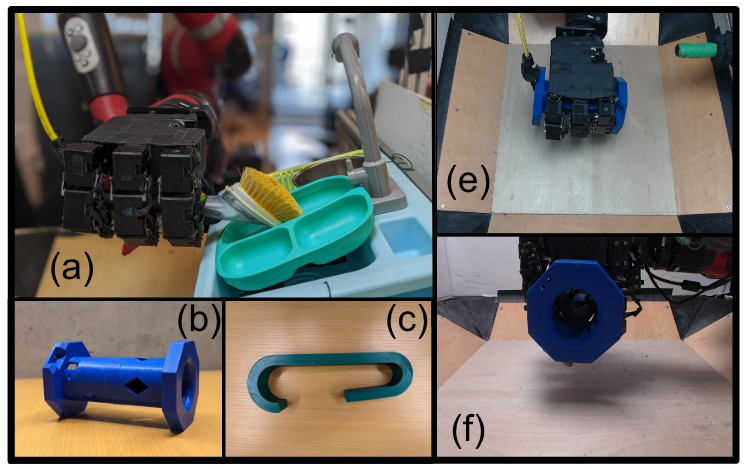}
  \caption{An overview of our experimental platform: (a) Our robot consists of a 16 DoF four-finger hand mounted on a 7 DoF Sawyer arm; (a, b, c) Objects the robot manipulates in our experiments: a kitchen brush, a cylindrical hose connector, and a hook that must be attached to a handle; (e, f) Observations for the robot come from two monocular RGB cameras.}
  \label{fig:hardware-setup}
  \vspace{-0.5cm}
\end{figure}

Our robotic system consists of a custom-built, 4-finger, 16-DoF robot hand, mounted on a 7-DoF Sawyer robotic arm. The arm and hand assembly are positioned over a tabletop surface (Fig.~\ref{fig:hardware-setup}, left image). Our policy, which we operate at 8Hz, directly controls each joint position in addition to the Cartesian position and orientation of the arm, resulting in a 22-dimensional action space and 29-dimensional state space. The system is designed to operate for upwards of 48 hours in contact-rich environments without breakage. In addition to the robot's own joint encoders, two RGB image observations are provided to the robot via two low-cost web cameras and resized to $84\times84$. We discuss additional details on our project website.

Our tasks consist of manipulation behaviors such as reaching, grasping, in-hand and mid-air reorienting, and inserting. We consider three tasks (shown in Fig~\ref{fig:hardware-setup}) for interacting with several different objects: inserting a hose into a connector on the side of the arena, hooking a rope onto a fixture, and cleaning a surface with a kitchen brush. Successfully completing each of these tasks requires correctly sequencing a series of sub-skills. For example, in order to complete our kitchen task, the robot must grasp and reorient that brush palm down without dropping it before making contact with the surface. Constructing and tuning both manual rewards and state estimation systems for each of these tasks separately would typically require laborious human engineering. For each of these tasks however, the only supervision we assume is to allow the user to place the robot and object in the desired position and capture a set of image ``snapshots''. We describe how we use this supervision to drive reward inference and task selection via autonomous reinforcement learning in the sections below.

\section{Problem Formalism and User Assumptions}

In this section, we formalize our problem setting and supervision assumptions. Consider first the Markov decision process (MDP) defined by the tuple $(\mathcal{S}, \mathcal{A}, p_{dyn}, \rho, \gamma, R)$, where $\mathcal{S}$ denotes the state space, $\mathcal{A}$ denotes the action spaces, $p_{dyn}: \mathcal{S} \times \mathcal{A} \times \mathcal{S} \mapsto \mathbb{R}_{\geq 0}$ denotes the environment dynamics, $R: \mathcal{S} \times \mathcal{A}\mapsto \mathbb{R}$ denotes the reward function, $\rho: \mathcal{S} \mapsto \mathbb{R}_{\geq 0}$ denotes the initial state distribution and $\gamma \in [0, 1)$ denotes the discount factor. The typical objective of episodic RL is to optimize the discounted return $J(\pi) = \mathbb{E}_{\tau \sim \pi} [\,\sum_{t=0}^{T} \gamma^t R(s_t, a_t)\,]$ with respect to the policy $\pi$, where $\tau = \{(s_i, a_i)\}_{i=0}^{T-1}$ is obtained by sampling ${s_0 \sim \rho(\cdot),  \textrm{ } a_t \sim \pi(\cdot \mid s_t) \textrm{ and } s_{t+1} \sim p( \cdot \mid s_{t}, a_{t})}$.

A principal concern of our work is to ask the question of how best to instantiate RL systems in the real world with minimal per-task engineering, instrumentation and intervention. Standard RL assumes a reward function $R$ that in practice must often be hand engineered and tuned per-task by a user. This challenge is particularly acute in the dexterous manipulation setting where the desired behavior can often itself be composed of a sequence of complex ``sub-tasks'' (e.g., grasping, re-orienting, etc) with different objects that would need to be instrumented separately. In addition, independent of being challenging to learn, these sub-tasks must be appropriately sequenced in order to complete the task but also to allow the agent to continue to practice in the event of failure. This necessitates the provision of more fine-grained guidance via user supervision, while carefully balancing the cost of providing such supervision. Furthermore, most RL algorithms assume that the environment is episodic and resets are provided for free. This is not true when considering large scale autonomous operation. 

To provide fine-grained supervision both for reward inference and autonomous practicing, we propose a method where a user supplies the robot with a set of sub-problems to practice. These sub-problems are defined by ``milestone'' examples, which constitute a graph structure:
\vspace{-0.1cm}
\begin{tcolorbox}
\vspace{-0.1cm}
\begin{definition}[\textit{Milestones graph}]
\label{def:graph}
We assume the user provides a set of outcome images that can be summarized by a graph $\mathcal{G} = (\mathcal{V}, \mathcal{E})$ of cardinality $|K|$ indexed by $z$, where each vertex $v \in \mathcal{V}$ is composed of a set of $M$ outcome images $\{s_{i}^z \}_{i=1}^{M}$.
Each set of outcome images characterizes a semantically meaningful sub-task to be solved. In addition, upon accomplishing a sub-task, a directed edge $(v, v') \in \mathcal{E}$, or equivalently a binary label, is provided which indicates which sub-task is to be practiced next.
\end{definition}
\vspace{-0.2cm}
\end{tcolorbox}
\vspace{-0.1cm}
Consistent with the goal of having the agent continuously practice (i.e., not get stuck), we assume there are no sink nodes in the provided graph.\footnote{This assumption conceivably could be lifted by providing handling of safety-critical states to avoid irreversible sinks,~\cite{garcia2015comprehensive,srinivasan2020learning}. For simplicity, we leave addressing safety issues for future work.} Then, instead of optimizing a single-task objective $J(\pi)$, we instead optimize all of the sub-tasks in the milestone graph simultaneously, resulting in a multi-task RL problem. Concretely, we learn a set of $K$ policies $\pi_z$ indexed by a categorical variable $z$ (one for each milestone), optimizing a set of MDPs, $\mathcal{M} \equiv (\mathcal{S}, \mathcal{A}, p_{dyn}(s_{t+1} | a_{t}, s_{t}), \{R_z\}_{z=0}^{K-1}, p_{\text{task}}(z | s))$ ,
where we have introduced a per milestone reward $R_z$ and task predictor $p_{\text{task}}$. This leads to the following objective:

\begin{align}\label{eq:multi-task-rl-train}
    J_{\text{MT}}(\{\pi_i\}_{i=0}^{K-1}) = &  \sum_{i=0}^\infty \left[\mathop{\mathbb{E}}_{\substack{\\ s_0^i = s_T^{i-1} \\ \tau \sim \pi_{z_i}}}\left[\sum_{t=0}^{T} \gamma^t R_{z}(s_t^i, a_t^i)\right]\right] \\
    & z_{i+1} \sim p_{\text{task}}(z_{i+1}| s_T^{i}).
\end{align}
Formulating the problem in this manner makes conspicuous the need to define and sample from $p_{\text{task}}(z_{i+1}| s_T^{i})$ and $R_{z}(s_t, a_t)$ which can be used to autonomously direct training. To tractably learn these two functions, we leverage the user-provided milestone supervision, which we denote as $D_k = \{s_i\}^{N}_{i=1}$, and a set of categorical labels $y_k$ from the milestones graph indicating which task should transition to the next. These could, for example, be a set of images showing the robot repositioning and re-grasping the object, and a label indicating the next step is pickup. In the following sections, we show how to learn the rewards $R_z$, the sub-task transition function $p_{\text{task}}(z_{i+1}| s_T^{i})$ and per sub-task policies $\pi_z$ using the milestone image supervision provided upfront. 

\begin{figure}[t]
    \centering
    \vspace{0.1in}
    \includegraphics[width=0.99\columnwidth]{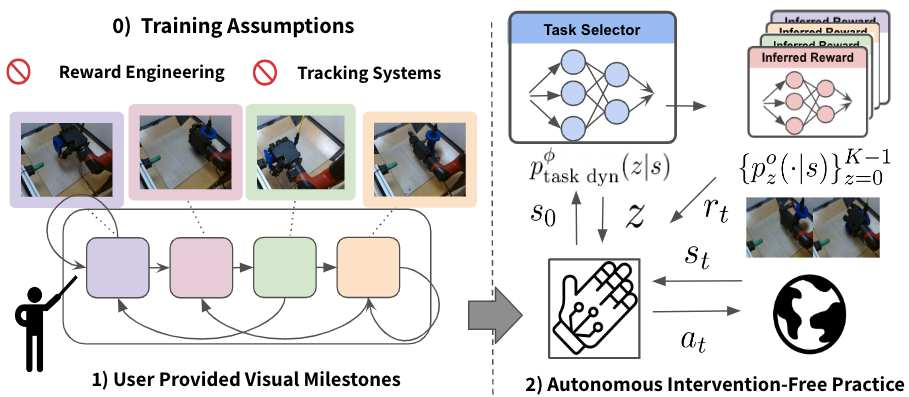}
    \caption{An overview of our approach. The user provides a set of visual milestones (bottom left, $K=4$ here) and transitions. We use this to formulate a multi-task learning problem where we leverage this supervision to learn all the components necessary for intervention-free learning. Prior to training, we learn a task selection model (blue module), which is used to choose amongst a set of $K$ policies for data collection (Sec.~\ref{sec:task_graph}). This data is used to learn a success classifier (top right, Sec.~\ref{sec:rew}), which is used to automatically assign rewards. We show that our system is capable of autonomously learning complex manipulation behaviors on a real world anthropomorphic hand with modest instrumentation beyond the robot's own joint encoders and camera.}
    \label{fig:system}
    \vspace{-0.2in}
\end{figure}

\section{The AVAIL System: Autonomy via User-Provided Milestones} 
\label{sec:method}

To address the problem described in Section~\ref{sec:prelim}, we present AVAIL (Autonomy ViA mILestones) -- our system for learning autonomously with minimal intervention and external environment instrumentation. AVAIL (see Fig.~\ref{fig:system} for an overview) reframes the RL training process as a multi-task problem that can learn directly from sparse milestone examples provided by users, with minimal external instrumentation or intervention. This makes the process of solving complex tasks with ``programming-free" reinforcement learning significantly more approachable. By leveraging the user-provided milestones defined in Sec~\ref{sec:prelim}, at a high level our system functions (as shown in Fig~\ref{fig:system}) by (1) deriving reward functions via learned success classifiers, (2) optimizing these rewards in a sample-efficient manner using a multi-task vision-based RL system, and (3) determining which of these tasks to perform given the current robot observations so as to continue practicing autonomously.

\subsection{Visual Multi-Task Policy and Reward Learning from User Milestones}
\label{sec:rew}

A critical enabler of autonomous learning is the ability for the robot to assign rewards to its own experience. This importantly relieves the burden of manual reward engineering, but comes with the trade-off of removing the ability of the designer to provide task information (e.g., via reward shaping), which can be crucial for tractable learning and directed exploration for long horizon tasks. 
To resolve this challenge in our setting where we similarly require compound behavior, we leverage the sparse milestone supervision to learn a set of success classifiers $\{\classifier_z(\cdot | s)\}_{z=0}^{K-1}$ that decomposes the entire long horizon task.
Importantly, the provision of a number of significant milestones takes the burden off a single learned classifier to provide accurate reward shaping.

For each individual classifier $\classifier_z(\cdot | s)$, we build on the VICE algorithm~\cite{fu2018variational}, extending it to a multi-task setting. We learn a binary classifier $\classifier_z(\cdot | s)$ over the set of user-provided examples $\{D_1, D_2, \dots, D_K\}$ as the positive class, and the agent's own experience sampled on-policy as the negative class. Once trained, the classifier probability $\classifier_z(o | s)$, or a monotonic transformation of it (e.g., $\log \classifier_z(\cdot | s)$), can be used as $R_z$ from Section~\ref{sec:prelim}. An added advantage of classifier-based rewards is the property that, in practice, they can often provide some additional degree of shaping~\cite{fu2018variational,mural}.

Finally, in order to learn in the real world from raw sensory inputs, a core component of our system is a sample-efficient, multi-task RL algorithm that allows us to learn the robot's joint encoders and raw image observations. In particular, given the set of reward functions discussed above, we learn a set of corresponding $K$ policies $\pi_z$ using the recently proposed DroQ approach~\cite{hiraoka2021dropout}. As an overview, DroQ is an approach which combines data augmentation in the form of random crops~\cite{kostrikov2020image} and implicit ensembling via dropout to allow for robust, sample-efficient image based learning. We refer the reader to Hiraoka et al.~\cite{hiraoka2021dropout} for a detailed description of DroQ and our project website for additional details on architecture choice and training hyperparameters.
 
\captionsetup[subfigure]{labelformat=empty}
\begin{figure*}[!t]%
    \centering
    \subfloat[]{{\includegraphics[width=0.99\columnwidth]{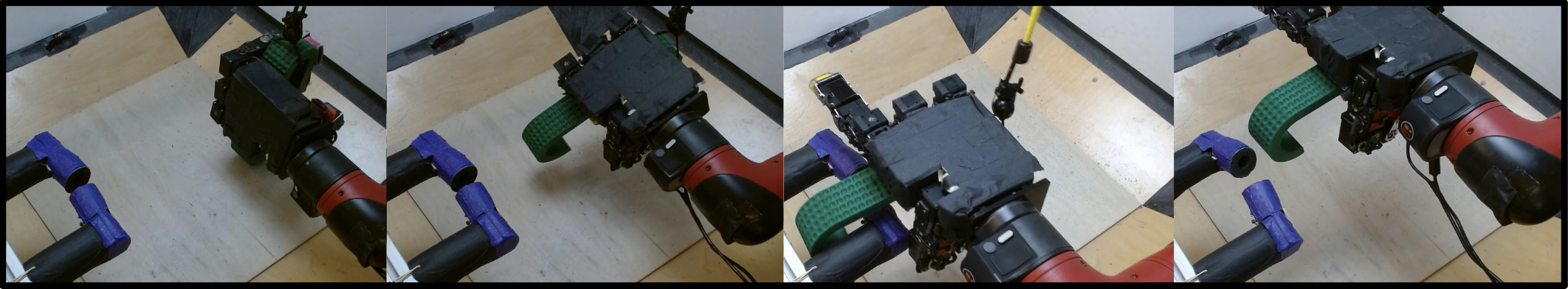}}}%
    \enspace
    \subfloat[]{{\includegraphics[width=0.98\columnwidth]{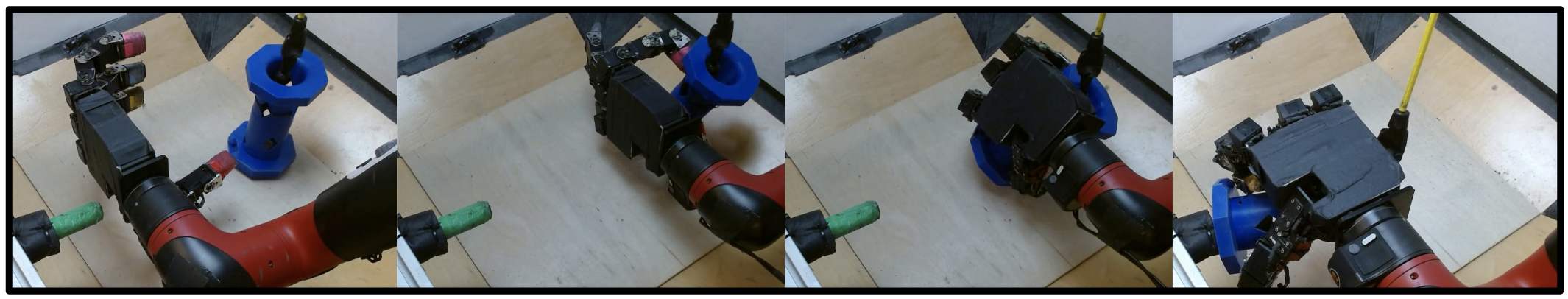}}}%
    \vspace{-0.2in}
    \caption{Filmstrip of the final learned hooking (left) and insertion behavior (right). Using the user-provided milestones, our robot learns a set of skills that allows it to autonomously practice hooking and unhooking (left, right two images) and recover from failure (e.g., after dropping the hook) by regrasping and reorienting the hook (left, left two images). Similarly, using the user-provided milestones, our robot learns a set of skills (e.g., grasp, insert), which together enable successful insertion (right, rightmost image) as well as the stages needed to practice autonomously (right, left three images). After 36 hours of unattended training, our system hooks onto the handle with around a 95\% success rate and successfully inserts with around 80\% success rate.}%
    \label{fig:filmstrip_hook_pipe}%
    \vspace{-0.3in}
\end{figure*}
\subsection{Multi-Task Learning without Oracles}
\label{sec:task_graph}

After attempting a particular task, the agent must decide which task to attempt next, which depends on the current situation (e.g., if it drops the hose connector, it should try to grasp it again, but if it is still holding it, it can attempt the insertion). In order to infer which task the agent should execute, we allow the user to provide next milestone labels (labels of what discrete milestone $z_k$ should be attempted at a particular state $s_k$) to train a task dynamics model by performing supervised learning over the milestones and labels provided at the beginning of training. That is, given a dataset of $\mathcal{D} = \{(z_k, s_k)_{k=0}^N\}$, we can recover 
$p_{\text{task}}^\phi = \argmax_{\phi} \mathbb{E}_{z_k, s_{k} \sim \mathbf{D}}\left[ \log  p_{\text{task}}^\phi(z_k | s_{k}) \right].$
These labels represent which task to perform upon success (e.g., the success examples for hooking the rope could be labeled with the `unhook' task).
During autonomous training, the agent samples a task from this learned model, which is then used to execute the corresponding policy $\pi_z$ in the environment. Rather than re-sample this task indicator at every step, we sample it every $T$ steps and keep it fixed during data collection. This scheme explicitly separates the task inference, and task learning allows each ``sub-problem'' to be treated effectively as a separate MDP.

\subsection{Algorithm Summary and Implementation Details}

To summarize, given the milestone graph provided by the user, our system, \methodname{} (Autonomy ViA mILestones), proceeds as follows. First, \methodname{}
performs supervised learning of the next task transitions provided by the user as described in Sec.~\ref{sec:task_graph} to learn $p_{\text{task}}^\phi(z | s)$. Next, during training, our approach chooses the most probable task $z$ using an observed state, which is then used to collect experience using the corresponding policy $\pi_z$. We train a set of separate policies $\pi_z$ for each of the $K$ sets of example images, with separate critics $Q_z$ and replay buffers $\mathcal{B}_z$. We parameterize each policy $\pi_z$ as a deep neural network, and train each policy using the soft actor-critic algorithm (SAC)~\cite{haarnoja2018soft} using rewards $R_z$ that are inferred via the multi-task VICE~\cite{fu2018variational} algorithm trained in the loop. Finally, rather than resampling the task every step, we do so every $T=100$ steps. 

\section{Experimental Evaluation}

Our experiments first aim to evaluate whether \methodname{} can learn complex manipulation skills in the real world with visually indicated milestones. To do so, we evaluate our approach on three real world manipulation tasks that require successfully sequencing a set of skills and performing complex coordinated finger motions to manipulate objects. We describe first our real world evaluation followed by our evaluation in simulation which we use to provide a rigorous comparison with prior methods. The results are best viewed in the supplementary videos provided on the project website: \href{https://sites.google.com/view/dexterous-avail/}{https://sites.google.com/view/dexterous-avail/}

\subsection{Real-World Task Descriptions}

We begin by describing our tasks. The objects and workspace in our experiments can be seen in Fig.~\ref{fig:hardware-setup}. The environments are mostly uninstrumented, except for a passive tether that prevents the object from falling out of reach. Further task details and demonstrations can be found on the project website.

\begin{figure*}[t]
\centering
\includegraphics[width=0.8\textwidth]{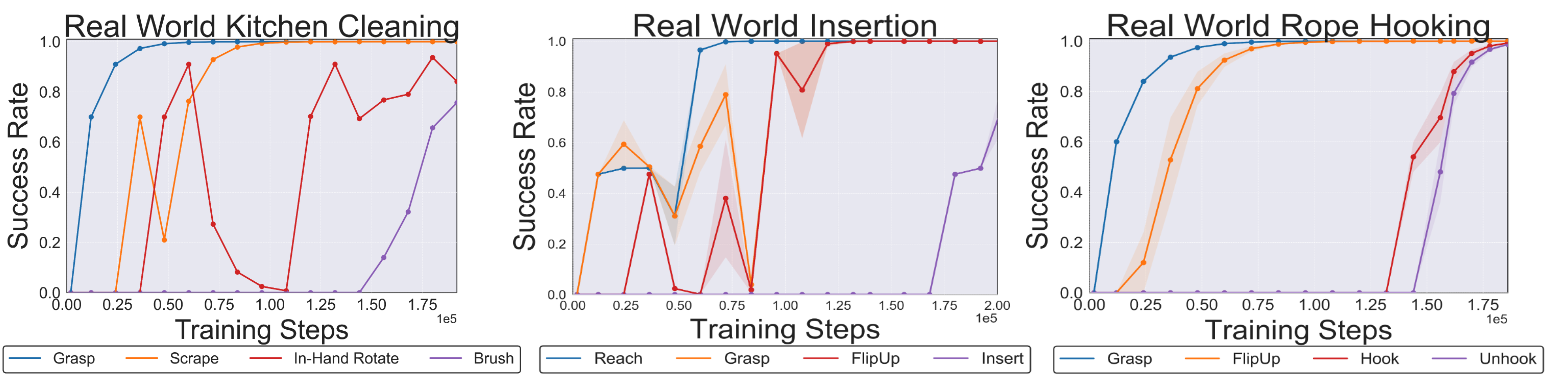}
\vspace{-0.05in}
\caption{Success rates of different milestones across our real world dexterous manipulation tasks. Our method is able to perform all the kitchen milestones with around 80\% success rate (left), successfully perform pipe insertion (middle, red curve) with around 80\% success, and nearly perfectly hook and unhook the rope (right, red/purple curve).
    Overall, success on other milestones improves earlier in training (blue, orange curves), equipping the robot with skills to autonomously retry the task.}
\label{fig:hardware_curves_main}
\vspace{-0.2in}
\end{figure*}

\noindent\paragraph{Using a kitchen brush.} The goal of the first task is to scrape a plate with a two sided cleaning brush (see Fig.~\ref{fig:teaser}). This task requires the robot to grasp the brush, and reorient it with the fingers so the bristles face the plate. The palm-down manipulation of the brush is challenging, as it requires balancing it so that it doesn't fall and rotating it around its long axis via a coordinated finger gait. The task milestones consist of grasping the brush, scraping the surface, reorienting it, and bringing the bristles in contact with the plate.

\noindent\paragraph{Hose connector insertion.} The goal of the second task is to attach a cylindrical hose connector to a peg connector, which requires reaching, grasping, reorienting, and performing the insertion as the task milestones. While the task is simpler in terms of dexterity, it requires visual perception to carefully insert the connector onto the peg connector (see Fig.~\ref{fig:filmstrip_hook_pipe}, left).

\noindent\paragraph{Rope hooking.} The third task requires attaching a hook to a handle (see Fig.~\ref{fig:filmstrip_hook_pipe}, right). The robot must grasp, reorient, and hook and unhook the object for its visual milestones. This task requires visually servoing the hook over a handle.

\subsection{Real-world evaluation}
To evaluate our system, we save the policies at regular intervals and evaluate their performance after training, so as not to interrupt the training process. For all tasks, the evaluation metric for each milestone is a binary success measurement based on the distance of the hand and object to the desired pose. We provide more details on our evaluation setup on our project website.

\noindent\textbf{Real-world skill learning:} We plot the performance of the evaluation runs as a function of the training step at which the policy was recorded in Fig.~\ref{fig:hardware_curves_main},
providing learning curves for real-world training. Observe that \methodname{} automatically provides a degree of scaffolding by successfully learning skills early on in training (blue curve in left plot, blue and orange curve in center right plot) that correspond to being able to regrasp and reorient the object. This allows the robot to continuously retry later milestones. By the end of training, we find that the robot is able to successfully perform all milestones with a $>80$\% success rate. We note that for all three tasks, no additional instrumentation is required beyond changing the object and fixture. Upon specifying the visual milestones, the robot is capable of completely unattended learning for approximately 48 hours of robot time. 

\begin{figure}[t]
    \centering
    \includegraphics[width=0.875\columnwidth]{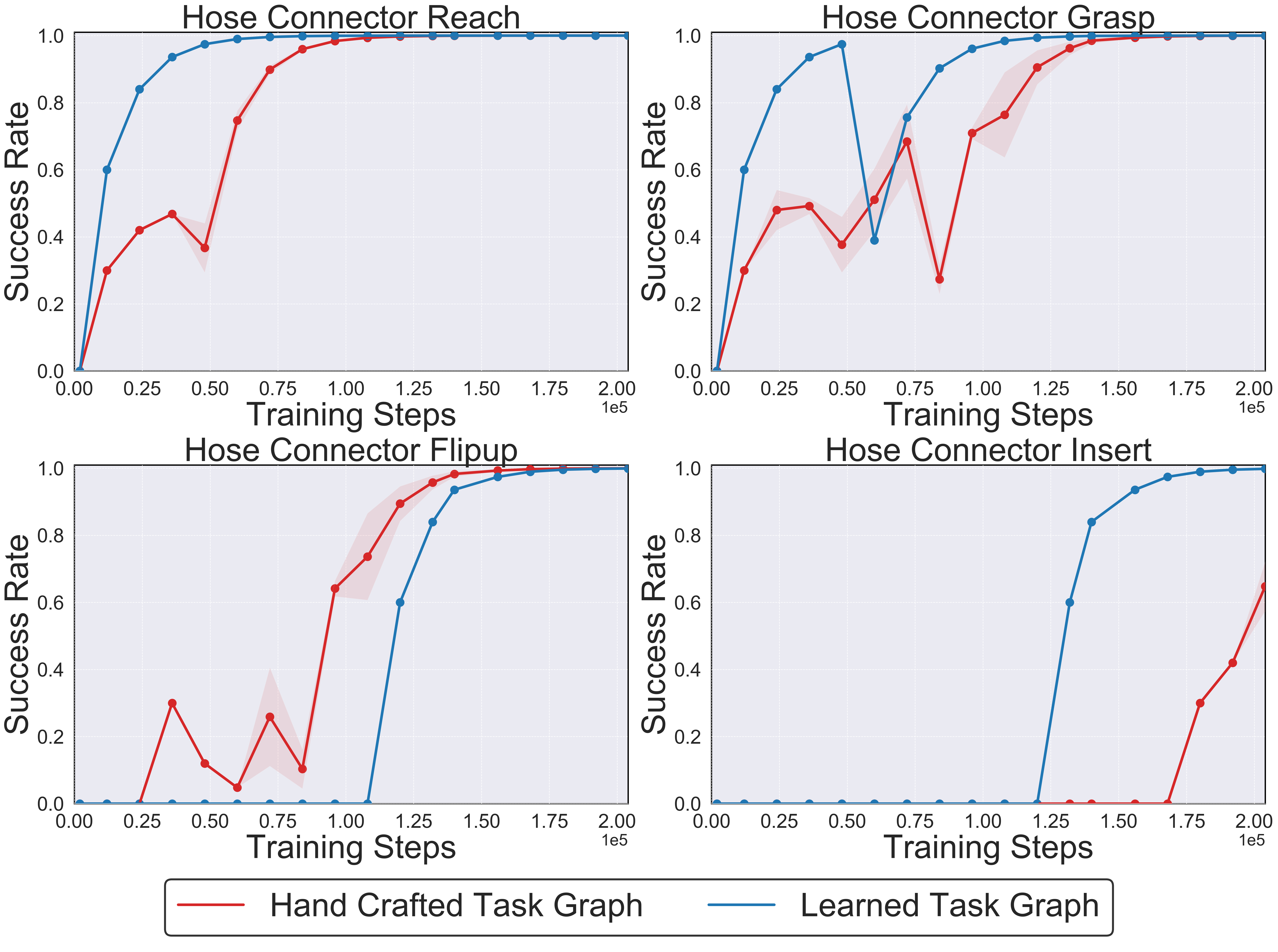}
    \caption{A comparison of the success rate of each task on our real world insertion task. We find that using a learned task graph results in faster convergence on our real world robotic task compared to a handcoded heuristic based task graph. We find the robot begins to consistently perform the final insertion task 25\% faster than a hand-coded task graph.}
    \label{fig:hardware_curves_pipe}
    \vspace{-0.3in}
\end{figure}

\noindent\textbf{Real-world task graph learning:} Next, to understand the effect of our learned task graph, we compare to a hand-designed task graph on our insertion task. This hand-designed task graph encodes a heuristic strategy where each of the tasks are practiced sequentially.
The comparative success rate can be seen in Fig.~\ref{fig:hardware_curves_pipe}. 
We find that our learned task graph outperforms this heuristic based task graph in terms of sample efficiency. We provide additional analysis on our real world rope hooking task on our project website, in addition to simulated analysis. Overall, we find that our approach is robust to errors in our learned task graph, although future work could likely improve overall sample efficiency by improving task graph training.

\subsection{Simulated Comparison}
Finally, we compare \methodname{} to prior autonomous RL method on the (\texttt{DHandValvePickup-v0}) simulation domain developed in prior work~\cite{gupta2021reset}. We first compare to a standard state-of-the-art RL algorithm, soft actor-critic~\cite{haarnoja2018soft} (which we denote as \textit{SAC}) using a reward learned from example images of a successful pickup or sparse reward. Note that prior work has used this approach for real-world robotics tasks~\cite{singh2019end}.
Next, we compare to a forward backward controller~\cite{eysenbach2017leave}, which can be seen as providing two milestones: one to pick up the object, and one to place it back on the table.
Finally, we compare to R3L~\cite{r3l}, where we provide the ``forward'' policy with a set of pickup goals
and follow Zhu et al.~\cite{r3l} by interleaving training of the forward policy with a ``perturbation controller'' trained with an intrinsic reward based on random network distillation~\cite{RND}. 

\begin{figure}[t]
  \centering
  \vspace{-0.15in}
  \includegraphics[width=0.7\columnwidth]{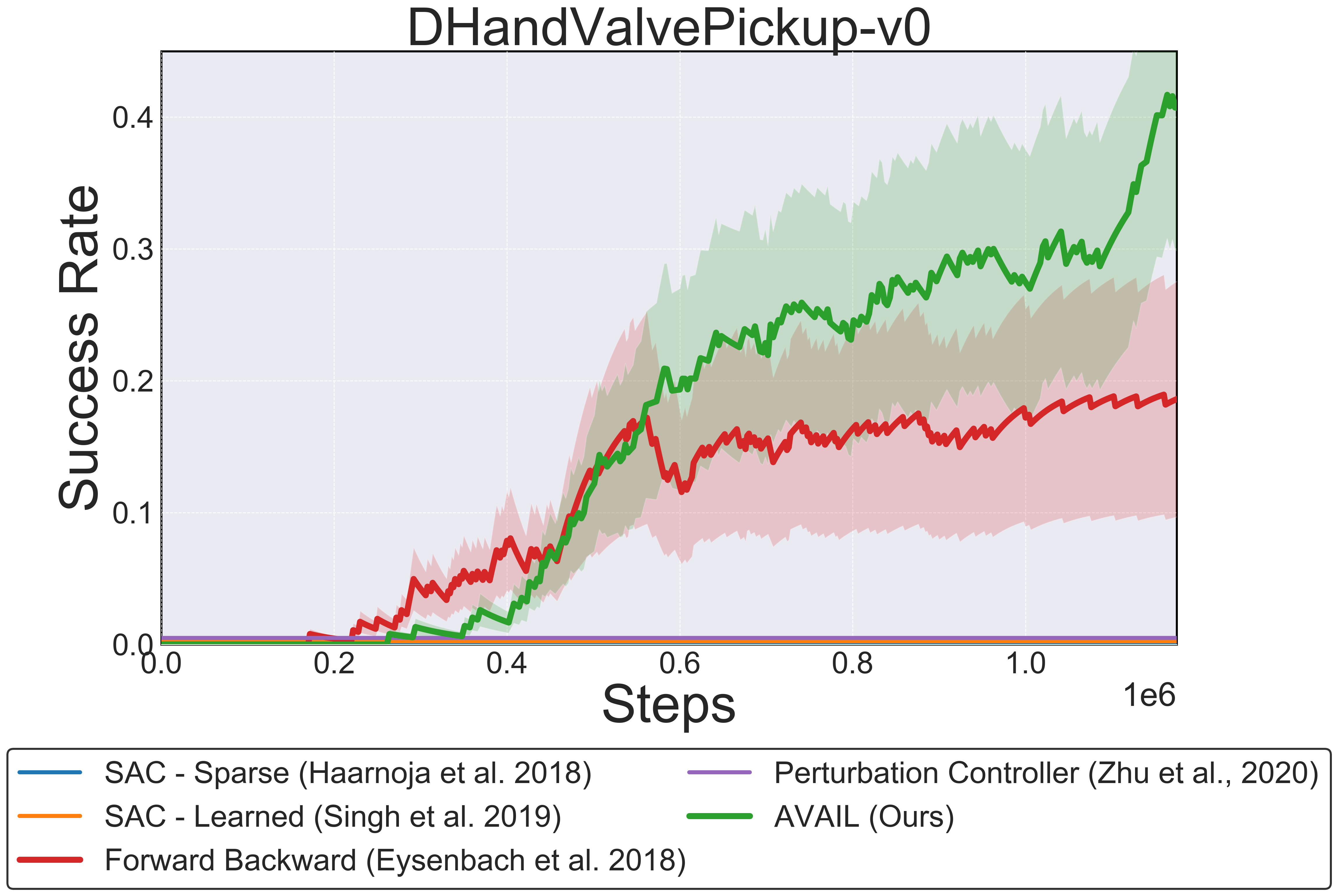}
  \vspace{-0.05in}
  \caption{\footnotesize{Success rates of each method averaged over 5 seeds for the full task on the \texttt{DHandValvePickup-v0} domain. Novelty based resets (purple) fail to make progress in this high DoF control problem. Compared to methods with fewer degree of supervision, $K=0,1,2$ milestones (blue, orange, red), our results illustrate the benefits of milestone supervision.}}
  \label{fig:evaluation_curves}
  \vspace{-0.15in}
\end{figure}

\noindent\textbf{Simulated comparative analysis.} We evaluate the performance of each method by sampling a random initial position of the object in the workspace and running the learned policy. We evaluate the task success over the course of training in Fig.~\ref{fig:evaluation_curves}.
Prior methods do not make successful progress on this task, due to the combination of reset-free training and the lack of a shaped reward. Without any handling of the reset-free setting, both variants of SAC fail to progress. While R3L in principle can handle the autonomous setting by perturbing the state between trials, the large, high-dimensional task simply provides too many ways for the purely novelty-seeking controller to modify the environment with a meaningful reset. The forward backward controller (red), which can be seen as an instantiation of our approach with two milestones, is the only prior method that succeeds in making progress.\footnote{We additionally compared the forward backward controller to our approach in the real world, but found it did not make progress on our tasks. We provide details of this analysis on our project website.}. Overall, this suggests that the improved performance can be achieved through providing more granular milestones.

\section{Discussion and Future Work}
We proposed a method for multi-task learning for dexterous manipulation from high dimensional image observations. Our method, \methodname, constructs a task graph from a modest number of user-provided milestone examples. This task graph illustrates how to practice and reset the task, and provides guidance to the learning process in lieu of more standard manual reward shaping. While the milestone examples require human effort to provide, we expect in many cases that this effort is significantly lower than providing full demonstrations. Much like a teacher or coach might instruct a student not just by telling them the \emph{goal} of a task but how they should go about practicing it, the milestone examples serve to provide guidance to the agent for how it should go about learning the desired behavior. Our experiments show that this approach effectively produces a learning process where the agent first practices the easier tasks, and then builds up the more complex tasks on top of them, all the while learning autonomously without resetting.

One limitation of our current system is that we do employ some physical instrumentation, by tethering the object so that it doesn't fall out of reach. We found that learning to pick up the object from any location was still too challenging for our method, because the range of possible situations was too large, and developing more capable RL methods that can address this is an important direction for future work. 

\section{Acknowledgement}
This research project was partially supported by the Office of Naval Research, with computing resources donated by Microsoft Azure.

\bibliographystyle{IEEEtran}
\bibliography{root}


\section*{Appendix}
\label{sec:appendix}

\section{Real World Environment Descriptions}

In the following sections, we describe the details of our real world tasks. We provide details related to experimental setup and describe our success criteria. Finally, we describe the supervision we provide the agent.

\subsection{Object, Arena Dimensions, and Safety Considerations}

\begin{figure}[H]
    \centering
    \subfloat[]{\label{fig:pipe}\includegraphics[width=0.40\columnwidth]{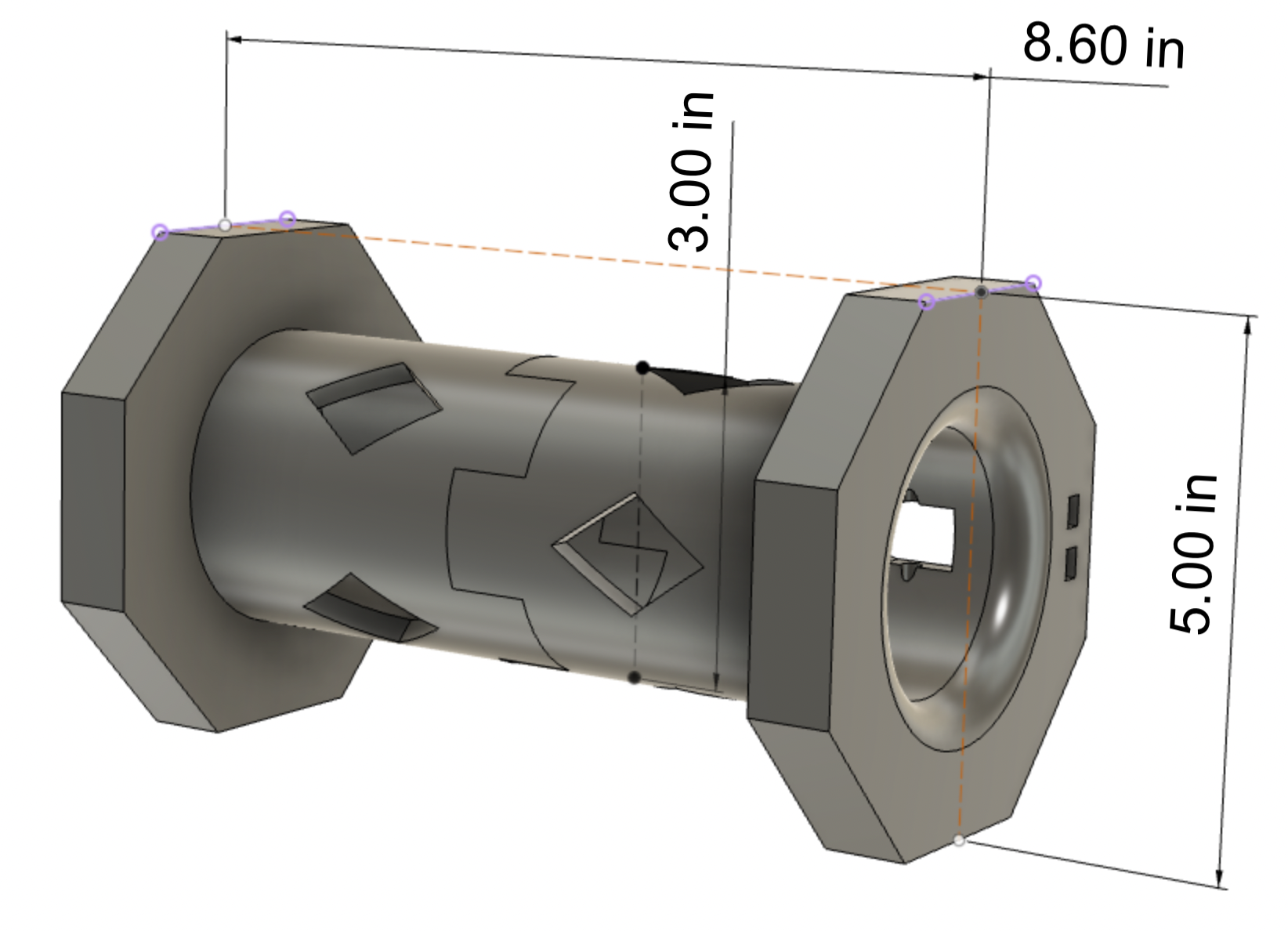}}
    \hfill
    \subfloat[]{\label{fig:hook}\includegraphics[width=0.40\columnwidth]{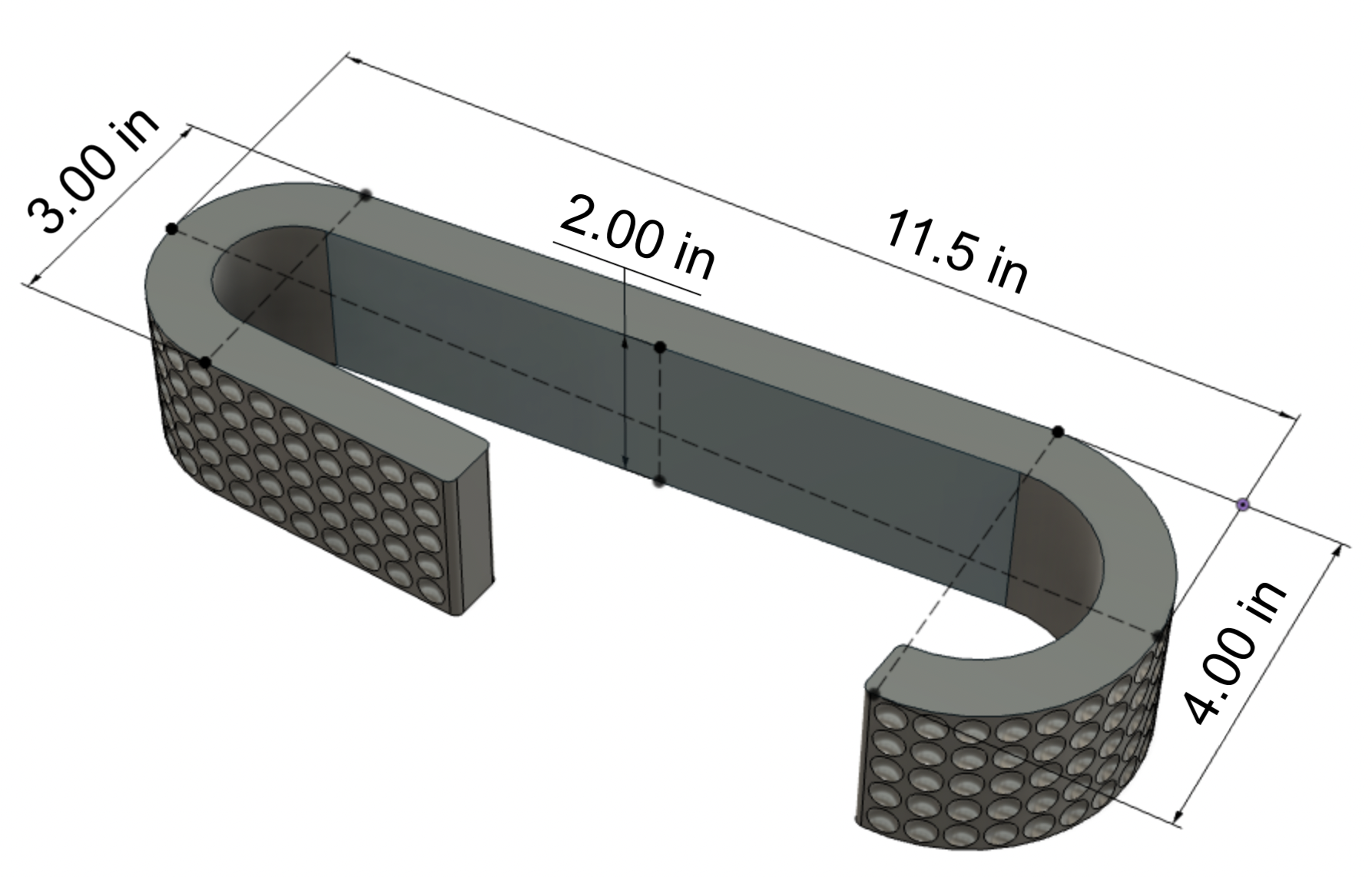}}
    \vspace{-0.2cm}
    \caption{\footnotesize{Technical drawing of objects (Left: hose connector, Right: hook) used for real-world experiments, with dimensions in inches.}}
\label{fig:hardware_drawings}
\vspace{-0.5cm}
\end{figure}

The objects used in our manipulation task are a 3-D printed pipe and hook that were custom designed. Their dimensions are shown in the technical drawing in Fig.~\ref{fig:hardware_drawings}. All the objects are manipulated in an arena of overall size 33" $\times$ 33" consisting of a base of 20" $\times$ 20" and 8" $\times$ 8" panels. Importantly, the fixtures we use, which can be seen in the example milestone images below, are made of a flexible foam. This is in order to ensure that the robot does not place excessive forces on either the object, hand, or fixture. We leave addressing these safety considerations to future work. 

\subsection{Hose Connector Insertion Task \& Evaluation Criteria}
\label{sec:appendix_hose}

The goal of this task is to have the hand reach, grasp, reorient, and insert a hose connector into an insertion point. The hose connector is attached to a fixed point at the top of the $20$in $\times$ $20$in arena by a rope that is $31$cm long. 

For this environment, we collected 300 milestone images ($84 \times 84 \times 6$) for each subtask. When training the VICE classifiers, we apply data augmentation using a random crop on the provided milestone images in addition to a randomly sampled $\mathcal{N}(0, 0.02)$ noise vector on the state.
\setlength{\fboxrule}{3pt}
\setlength{\fboxsep}{0pt}
\begin{figure}[h]
    \centering
    \fbox{\includegraphics[width=.8\columnwidth]{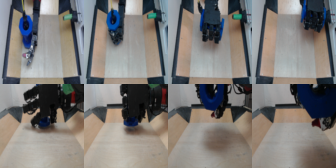}}
    \caption{\footnotesize{Sample milestone images for Reach, Grasp, Flipup, Insert (each column respectively))}}
    \label{fig:pipe_goal_images}
\end{figure}

In the following table, we summarize the success criteria that we use in our experiments.

\vspace{-0.25cm}
\begin{table}[H]
  \footnotesize
  \label{table:dclawhyperparams}
  \begin{center}
    \begin{tabular}{@{}ll@{}}
      \toprule
      \textbf{Task} & \textbf{Success Criteria}\\
      \midrule
      Reach & $\mathbbm{1}$ \Big\{$|x_{palm} - x_{hose}| < 0.05 $ AND $ |y_{palm} - y_{hose}| < 0.01$\Big\} \\
      \midrule
      Grasp & $\mathbbm{1}$ \Big\{ $is\_held\_during\_flipup$
      \Big\} \\
      \midrule
      FlipUp & $\mathbbm{1}$ \Big\{$|\theta_{hose} - \theta_{goal}| \le 5^{\circ}$\Big\} \\
      \midrule
      Insert & $\mathbbm{1}$ \Big\{$|x_{hose} - x_{peg}| < 0.05 $ AND $ |y_{hose} - y_{peg}| < 0.03$\Big\} \\
      \bottomrule
    \end{tabular}
  \end{center}
\end{table}
\vspace{-0.3cm}
In this environment, the evaluation criteria for successful finger grasps is intuitively defined as whether or not the grasp is firm enough for performing
subsequent tasks, i.e. the hose connector does not fall from the hand. $\theta_{hose}$ is the Euler angle measurement of the hose connector, where the optimal insertion angle is $\theta_{goal} = 90^{\circ}$. $x_{hose}$, ${x_{peg}}$, $y_{hose}$, ${y_{peg}}$ are the center of mass $xy$-coordinate of the hose connector and the insertion peg, measured in meters.

\subsection{Rope Hooking Task \& Evaluation Criteria}
\label{sec:appendix_rope}

The goal of this task is to have the hand grasp, reorient, hook, and unhook a carabiner hook onto a latch. The hook is attached to a fixed point at the top of the arena by a rope that is $31$cm long. 

For this environment, we also collect 300 milestone images ($84 \times 84 \times 6$) for each subtask. Similar to the insertion task, when training the VICE classifiers, a random crop is used as data augmentation for the milestone images and a random $\mathcal{N}(0, 0.02)$ noise is added to the milestone states.

\vspace{-0.25cm}
\begin{table}[h]
  \footnotesize
  \label{table:dclawhyperparams}
  \begin{center}
    \begin{tabular}{@{}ll@{}}
      \toprule
      \textbf{Task} & \textbf{Success Criteria}\\
      \midrule
      Grasp & $\mathbbm{1}$ \Big\{ $is\_held\_during\_flipup$ \Big\} \\
      \midrule
      FlipUp & $\mathbbm{1}$ \Big\{$|\theta_{hook} - \theta_{goal}| \le 5^{\circ}$\Big\} \\
      \midrule
      Hook & $\mathbbm{1}$ \Big\{$(x_{hook} - x_{latch}) > 0.01$\Big\} \\
      \midrule
      Remove & $\mathbbm{1}$ \Big\{$(x_{hook} - x_{latch}) < 0.05$\Big\} \\
      \bottomrule
    \end{tabular}
  \end{center}
\end{table}
\vspace{-0.3cm}

Similar to the hose insertion task, in this environment, the evaluation criteria for successful finger grasps is intuitively defined as whether the grasp is firm enough for performing
subsequent tasks, i.e. the hook object does not fall from the hand. $\theta_{hook}$ is the Euler angle measurement of the hook object along the side of its flat handle, where the ready-to-hook angle is $\theta_{goal} = 90^{\circ}$. $x_{hook}$ and ${x_{latch}}$ is the center of mass $x$-coordinate of the hook object and the latching bar, measured in meters.

\begin{figure}[h]
    \centering
    \fbox{\includegraphics[width=0.8\columnwidth]{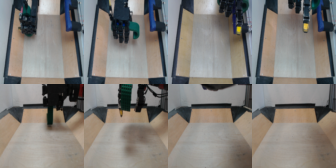}}
    \caption{\footnotesize{Sample milestone images for Grasp, Flipup, Hook, Unhook task, (each column respectively)}}
    \label{fig:hook_goal_images}
\end{figure}

\subsection{Kitchen Cleaning Task \& Evaluation Criteria}
\label{sec:appendix_kitchen}
This cleaning environment demonstrates our method and the robot's ability to perform challenging tasks in a real-world kitchen setup. The robot needs to grasp the dish-washing brush, scrape the plate with the plastic end, rotate the brush 180 degrees in the hand with the palm facing downward, and clean the plate. The brush is attached to a fixed point at the top of the arena by a rope that is $31$cm long. We use the same reward learning configurations as in the other two experiments.

\vspace{-0.25cm}
\begin{table}[h]
  \footnotesize
  \label{table:dclawhyperparams}
  \begin{center}
    \begin{tabular}{@{}ll@{}}
      \toprule
      \textbf{Task} & \textbf{Success Criteria}\\
      \midrule
      Grasp & $\mathbbm{1}$ \Big\{ $is\_held\_during\_flipup$ \Big\} \\
      \midrule
      Scrape & $\mathbbm{1}$ \Big\{$z_{scraper} - z_{plate} = 0$\Big\} \\
      \midrule
      Rotate & $\mathbbm{1}$ \Big\{$165^{\circ} \leq |\theta_{start} - \theta_{end}| \leq 195^{\circ}$\Big\} \\
      \midrule
      Brush & $\mathbbm{1}$ \Big\{$(z_{bristle} - z_{plate}) = 0$\Big\} \\
      \bottomrule
    \end{tabular}
  \end{center}
\end{table}
\vspace{-0.3cm}

In this environment, the success criteria for grasping the dish-washing brush is determined by whether the brush handle stays in hand when performing
subsequent tasks. $\theta_{start}$ and $\theta_{end}$ are the Euler angle measurements of the brush object along its longitudinal (roll) axis before and after the in-hand rotation task. The goal is to rotate the side with the bristle from pointing upward to facing the plate. $z_{scraper}$ and $z_{bristle}$ are the z-coordinates of the plastic scraper and the bristle on the brush. The goal for both the scrape and brush tasks is to apply these functional sites onto the surface of the plate.

\begin{figure}[h]
    \centering
    \includegraphics[width=1.0\columnwidth]{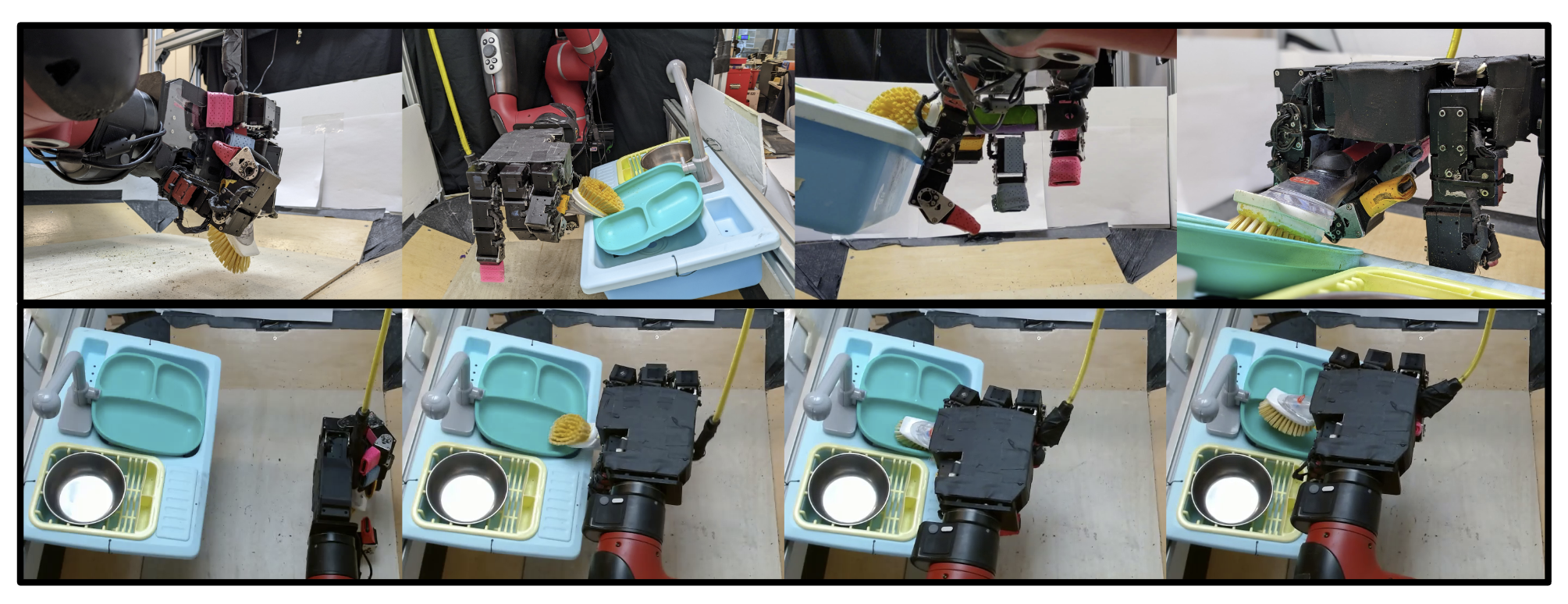}
    \caption{\footnotesize{Sample images for successful Grasp, Scrape, Rotate, Brush task, (each column respectively)}}
    \label{fig:brush_goal_images}
\end{figure}

\section{Simulated Environment Descriptions}
\label{sec:appendix_env}

We first describe the details of simulated environments used in the paper, and next list the hyperparameters used to train all the agents. The environment we use is based on the MuJoCo 2.0 simulator.

\subsection{Valve3 Task}
\label{appendix:ant}

The \texttt{DHandValve3} environment contains a green, three-pronged valve placed on top of a square arena with dimensions 0.55m x 0.55m. The valve contains a circular center, and each of its prongs has an equal length of 0.1m. There are three phases in this task: reach, reposition, and pickup.The reach phase's success criteria is when the hand is within 0.1m from the valve. In the reposition phase, the success criteria for the hand is to reach for the valve, grasp the valve with its fingers, and drag the valve to within 0.1m of the arena center. Finally, success of the pickup phase is measured if the object is picked up and brought within 0.1 m of the target location which is 0.2m above the table. In order to prevent the object from falling off the table, the object is constrained to a 0.15m radius by a string from the center of the table. 

The observation space of the environment is two camera views of the robot, resized to $84\times 84 \times 3$. In addition, the proprioceptive state of the arm is provided, which is comprised of a 16-dim hand joint position, 7-dim Sawyer arm state, and 6-dim vector representing the end-effector position and euler angle. The time horizon we use in each phase of the environment is $T=100$. We provide $300$ goal images per phase, which is comparable to the number used in prior work~\cite{fu2018variational}.

Below, we provide the oracle task graph for the valve environment, which we use to evaluate the relative performance of our learned task graph model.

\begin{algorithm}[h]
\begin{algorithmic}[1]
\REQUIRE Object position $\begin{bmatrix}x \\ y \\ z\end{bmatrix}$, previous task $\phi$
\STATE Let $\begin{bmatrix}x_{center} \\ y_{center}\end{bmatrix}$ be the center coordinates of the arena (relative to the Sawyer base).
\STATE Let $\begin{bmatrix}x_{hand} \\ y_{hand}\end{bmatrix}$ be the location coordinates of the hand (relative to the arena).
\STATE $is\_centered = ||\begin{bmatrix}x \\ y\end{bmatrix} - \begin{bmatrix}x_{center} \\ y_{center}\end{bmatrix}|| < 0.1$
\STATE $is\_hand\_over\_object = ||\begin{bmatrix}x \\ y \end{bmatrix} - \begin{bmatrix}x\_hand \\ y\_hand \end{bmatrix}|| < 0.15$
\IF{NOT $is\_centered$ and $is\_hand\_over\_object$}
\STATE Reposition
\ELSIF{NOT $is\_centered$ and NOT $is\_hand\_over\_object$}
\STATE Reach
\ELSIF{$is\_centered$}
\STATE Pickup
\ENDIF
\end{algorithmic}

\caption{{\bf Valve3 Task Graph (Oracle Baseline)}}
\label{alg:task_graph_3}
\end{algorithm}

\section{Algorithm Details}
\label{sec:appendix_methods}

In this section, we describe details related to our RL learning algorithms and also provide hyperparameters for each method.  

\subsection{AVAIL}
\label{sec:appendix_algo}
\begin{algorithm}[h]
 	\caption{\methodname{} (Autonomy ViA mILestones)}
 	\label{alg:method}
 	\begin{algorithmic}[1]
 	\STATE Given: $K$ tasks with examples states $\mathbf{D}:=\{D_z, y_z\}_{z=0}^{K-1}$, start state $s_0$.
 	\STATE Train task graph $p_{task dyn}(z | s)$ using $\mathbf{D}$  
 	\STATE Initialize $\pi_z$, $\classifier_z(o | s)$, $Q_z$, $\mathcal{B}_z$ for $z \in \{0, 1, \ldots, K - 1\}$
 	\FOR{iteration $n=1, 2, ...$}
 	    \STATE Select current task $z$ by querying learned task graph at the current state: $z = \text{argmax}_z \taskgraph$
        \FOR{iteration $j=1, 2, ..., T$}
            \STATE Execute $\pi_z$ in environment, storing data in the buffer $\mathcal{B}_z$
            \STATE Update the current task's policy and value functions  $\pi_z$, $Q_z$ using samples from $\mathcal{B}_z$, assigning reward based on $\classifier_z(o | s)$ using SAC~\cite{haarnoja2018soft}. 
            \STATE Update the classifier parameters, using $D_z$ and samples from $\mathcal{B}_z$ , using the VICE~\cite{fu2018variational}.
        \ENDFOR
 	\ENDFOR
 	\end{algorithmic}
\end{algorithm}

\subsection{Reinforcement Learning from Images}

For completeness, here we describe our procedure of performing image based RL, which, as noted in prior work, presents significant optimization challenges~\cite{laskin2020reinforcement, kostrikov2020image}.
In order to make learning more practical, we make use of a combination of data augmentation techniques during training, which has been previously shown to improve image based reinforcement learning~\cite{kostrikov2020image}, and dropout regularization~\cite{hiraoka2021dropout}. For all approaches we evaluate on in this work we make use of random shifts perturbations, which pad the image observation with boundary pixels before taking a random crop. \\
\\
We denote $s_{aug} \sim f(s)$ as an randomly augmented image from a distribution $f$. We compute Q-Learning by computing the Q value for a state ($s_i$) over $M$ independent augmentation. For each q function, $Q_\theta(s,a)$, we follow \cite{hiraoka2021dropout} and apply dropout followed by layer normalization in the fully connected layers of the critic.
\begin{align*}
\E_{\substack{s_{i} \sim B\\a \sim \pi(\cdot | s)}}[Q_\theta(s,a)] &\approx \frac{1}{M} \sum_{m=1}^{M} Q_\theta(f(s_{i}), a_m) \\ 
&\mbox{ where } a_m \sim \pi(\cdot|f(s_{i})),
\end{align*}
and computing a target value over $L$ augmentations 
\begin{align}
    y_i& = r_i + \gamma \cfrac{1}{L}\sum_{l=1}^{L} Q_{\theta}(f(s'_i, \nu'_{i,l}), a'_{i, l}) \\
    &\mbox { where } a'_{i, l} \sim \pi(\cdot | f(s'_i, \nu'_{i, l})).
    \label{eqn:y_target}
\end{align}
This leads to a final learning rule
\begin{align}
\theta &\leftarrow \theta - \lambda_{\theta} \nabla_\theta \frac{1}{N} \sum_{i=1}^{N} (Q_{\theta}(f(s_i, \nu_i), a_i) - y_i)^2.
\label{eqn:targetq_update}
\end{align}

\subsection{Hyperparameters}
Here we describe the individual prior methods we compare to in detail for the purpose of reproducibility. For shared parameters, we summarize them below and provide baseline specific parameters separately.

For the simulated experiments, we train a classifier with an identical architecture to our success classifiers. During training we sample a new task a horizon of $T=100$, as we found that in simulation the simpler na\"{i}ve task graph performs comparably to the learned task graph, in real world training we employ the na\"{i}ve task graph as it is arguably simpler. We provide additional experiments in the following section using the learned task which demonstrates that the learned task graph actually outperforms the na\"{i}ve task graph on the more challenging real world domains.

\vspace{-0.25cm}
\begin{table}[H]
  \footnotesize
  \label{table:global}
  \begin{center}
    \begin{tabular}{@{}ll@{}}
      \toprule
      \textbf{Shared RL Hyperparameters} & \textbf{Value}\\
      \midrule
      Base Encoder & Conv(3, 3, 32, 2)\\
                   & 3 $\times$  Conv(3, 3, 32, 1)\\
      Actor Architecture & FC(256, 256)\\
                          & FC(256, 22)\\
      Critic Architecture & FC(256, 256)\\
                          & FC(256, 1)\\
      \midrule
      Optimizer & Adam\\
      Learning rate & \{3e-4\}\\
      \midrule
      Discount $\gamma$ & 0.99 \\ 
      \midrule
      Target Update Frequency & 1 \\
      Actor Update Frequency & 1 \\ 
      \midrule
      Batch size & 256 \\
      Classifier batch size & 256 \\
      \bottomrule
    \end{tabular}
  \end{center}
\end{table}
\vspace{-0.3cm}

\vspace{-0.25cm}
\begin{table}[H]
  \footnotesize
  \label{table:vice}
  \begin{center}
    \begin{tabular}{@{}ll@{}}
      \toprule
      \textbf{Shared Classifier} \\ \textbf{Hyperparameters} & \textbf{Value}\\
      \midrule
      Optimizer & Adam\\
      Learning rate & \{3e-4\}\\
      Classifier steps per iteration & 1\\
      Mixup Augmentation $\alpha$ & 10\\
      Label Smoothing $\alpha$ & 0.1 \\ 
      \midrule
      Classifier Architecture & Conv(3, 3, 32, 2)\\
                              & 3 $\times$ Conv(3, 3, 32, 1)\\
                              & 3 $\times$ FC(512) $\rightarrow$ ReLU() $\rightarrow$ Dropout(0.5) \\
                              & FC(1) \\
      \bottomrule
    \end{tabular}
  \end{center}
\end{table}
\vspace{-0.3cm}

\section{Additional Real World Comparisons}

We additionally provide real world comparisons that mirror the more extensive comparisons done in simulation. When compared to prior methods, we find that our simulated results are corroborated by our real world experiments. We find that our approach outperforms the forward backward algorithm on our hose insertion and hook tasks. In evaluating the efficacy of our learned task, we find, however, that on the more challenging real world tasks, our method substantially performs a na\'{i}ve task graph. We provide additional details here.

\subsection{Real World Comparisons to the Forward-Backward Controller~\cite{eysenbach2017leave}}

In real world setting, we experiment and evaluate both our method and the Forward Backward (Eysenbach et al. 2018) method. For the Forward Backward method, we remove the training of the reach task by scripting it, making the learning problem easier. Additionally, we script the reach subtask, combine grasp and flipup into one forward task with horizon $T=100$ (50 for each task) and 600 milestone images, and train insertion as the backward task. Even when making the learning problem easier for the Forward Backward method setting by scripting the reach subtask, our method outperforms the Forward Backward method. The discrepancies in the performance of the algorithms demonstrate the improvements in learning capacity as the granularity of the division of tasks increases.

\begin{figure}[h]
    \centering
    \includegraphics[width=1.0\columnwidth]{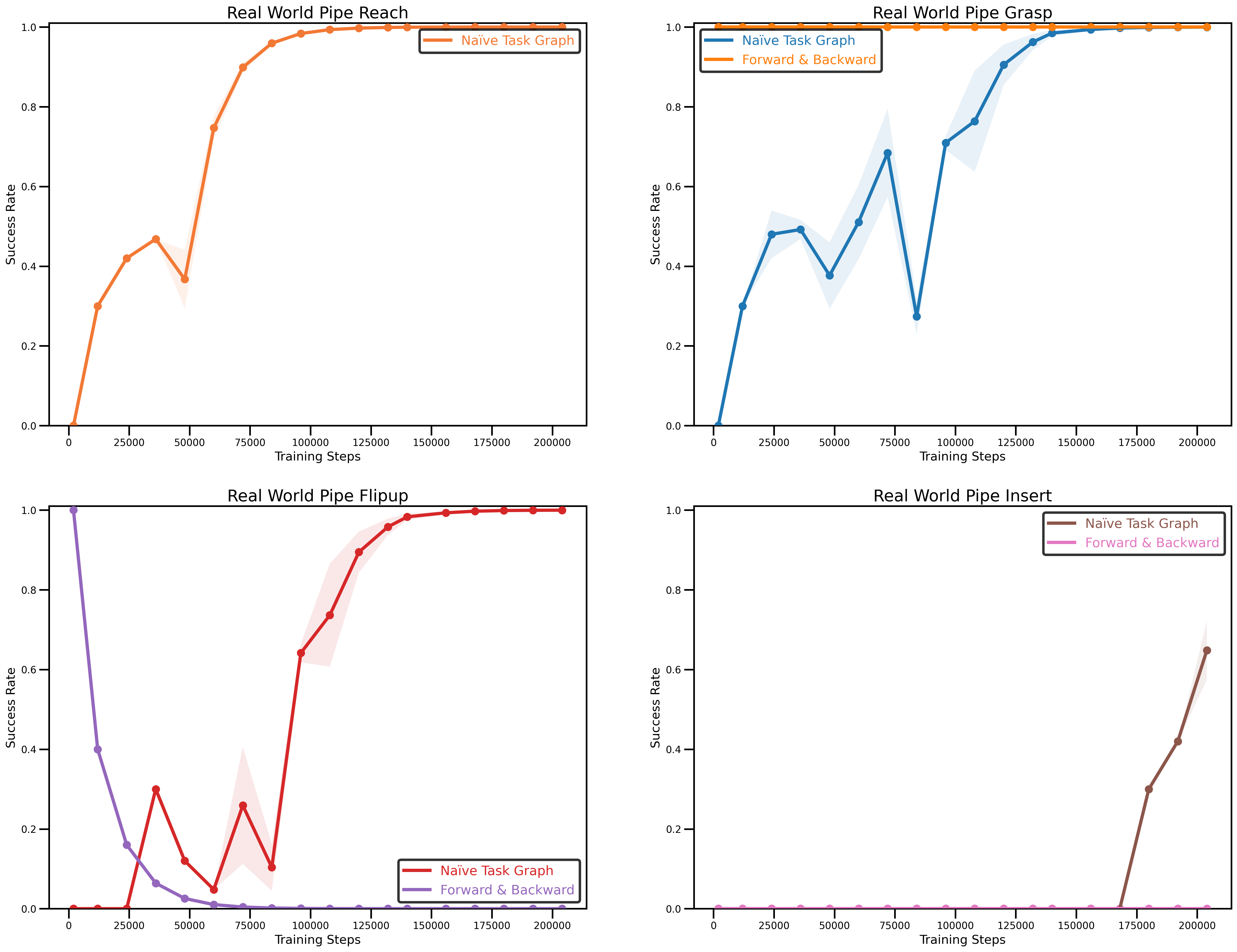}
    \vspace{-0.2cm}
    \caption{\footnotesize{Success rate of each subtask on our real world insertion task. The reach task curve with Forward Backward method is omitted as mentioned above. The Forward Backward method is unable to learn the flipup and insertion task while our method with the na\"{i}ve task graph achieves substantial learning progress across all tasks.}}
\label{fig:hardware_curves}
\end{figure}

For the Forward Backward method, we combine grasp, flipup, and hook into one forward task with horizon $T=200$ (50 for grasp and flipup, 100 for hook) and 900 milestone images, and train unhook as the backward task. Once again, since our method outperforms the Forward Backward method, the results highlight the improvements in learning capacity as the granularity of the division of tasks increases.

\begin{figure}[h]
    \centering
    \includegraphics[width=1.0\columnwidth]{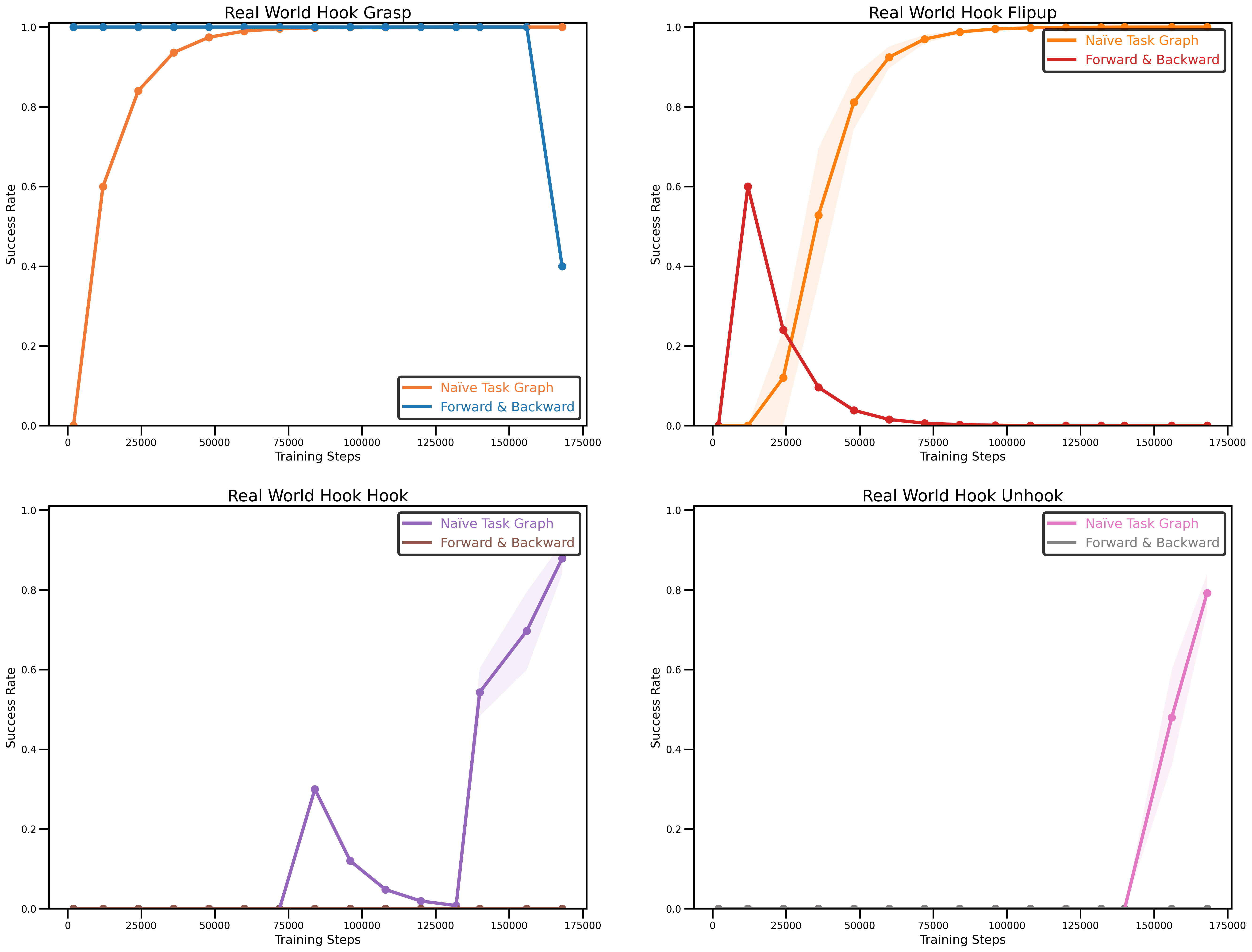}
    \vspace{-0.2cm}
    \caption{\footnotesize{Success rate of each task on our real world hooking task. The Forward Backward method is unable to learn the flipup, hook, and unhooking subtasks while our method with the na\"{i}ve task graph achieves substantial learning progress across all tasks.}}
\label{fig:hardware_curves}
\end{figure}

\subsection{Comparison with Learned Task Graph}

Here we show the success rate of our approach using a learned task graph on our real world insertion task. Different from our simulated domain, we find that on our more challenging real world tasks, we obtain significantly faster convergence (approximately 150,000 steps vs. 200,000 steps) in terms of final insertion performance using our learned task graph.

\begin{figure}[h]
    \centering
    \includegraphics[width=1.0\columnwidth]{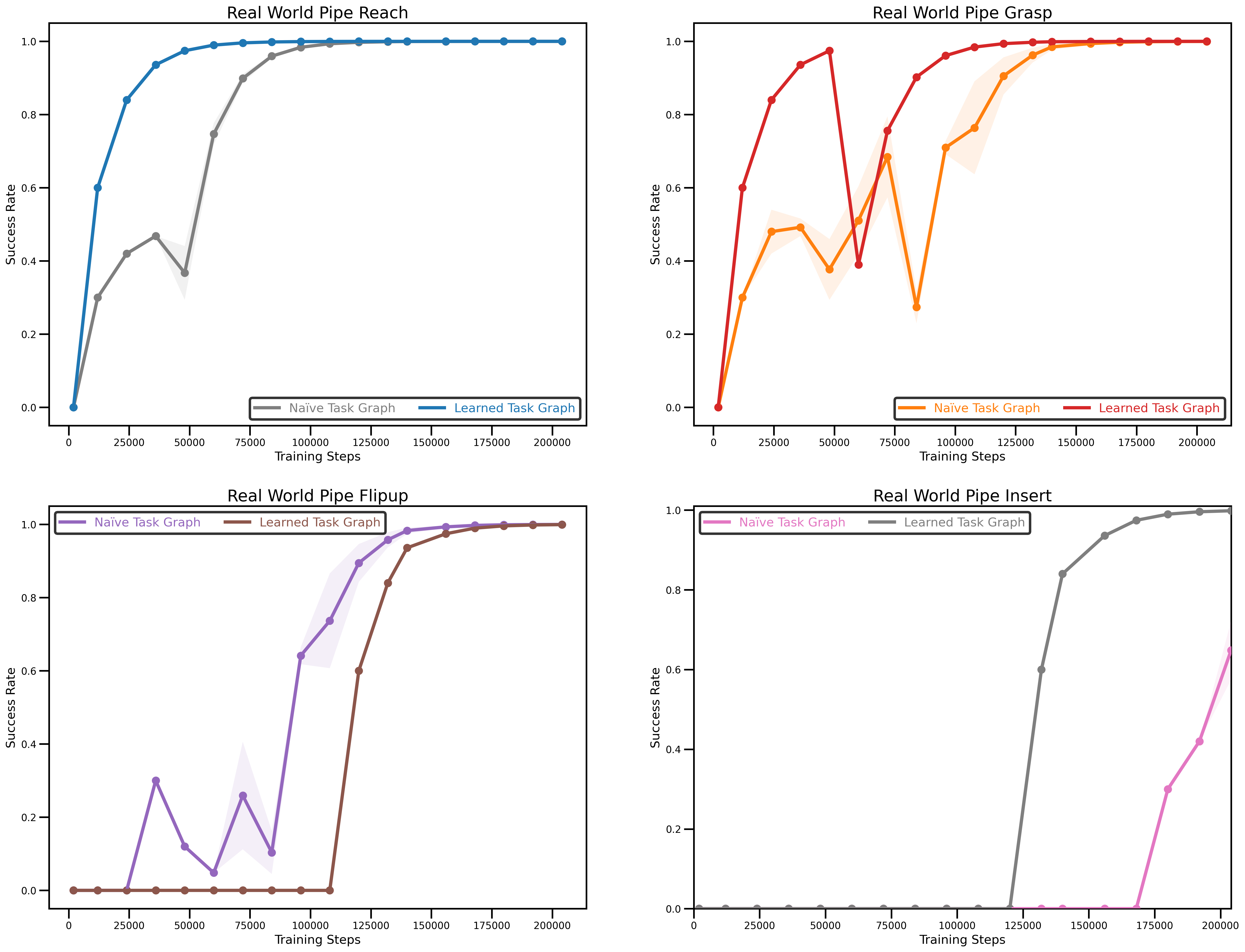}
    \vspace{-0.2cm}
    \caption{\footnotesize{Success rate of each task on our real world insertion task. We find that using a learned task graph results in faster convergence on our real world robotic task, where the robot begins to consistently perform the task around 25\% faster than the na\"{i}ve task graph.}}
\label{fig:hardware_curves}
\end{figure}

Here we show the success rate of our approach using a learned task graph on our real world hook task. Different from our simulated domain, we find that on our more challenging real world tasks, we obtain significantly faster convergence (approximately 130,000 steps vs. 160,000 steps) in terms of final insertion performance using our learned task graph.

\begin{figure}[h]
    \centering
    \includegraphics[width=1.0\columnwidth]{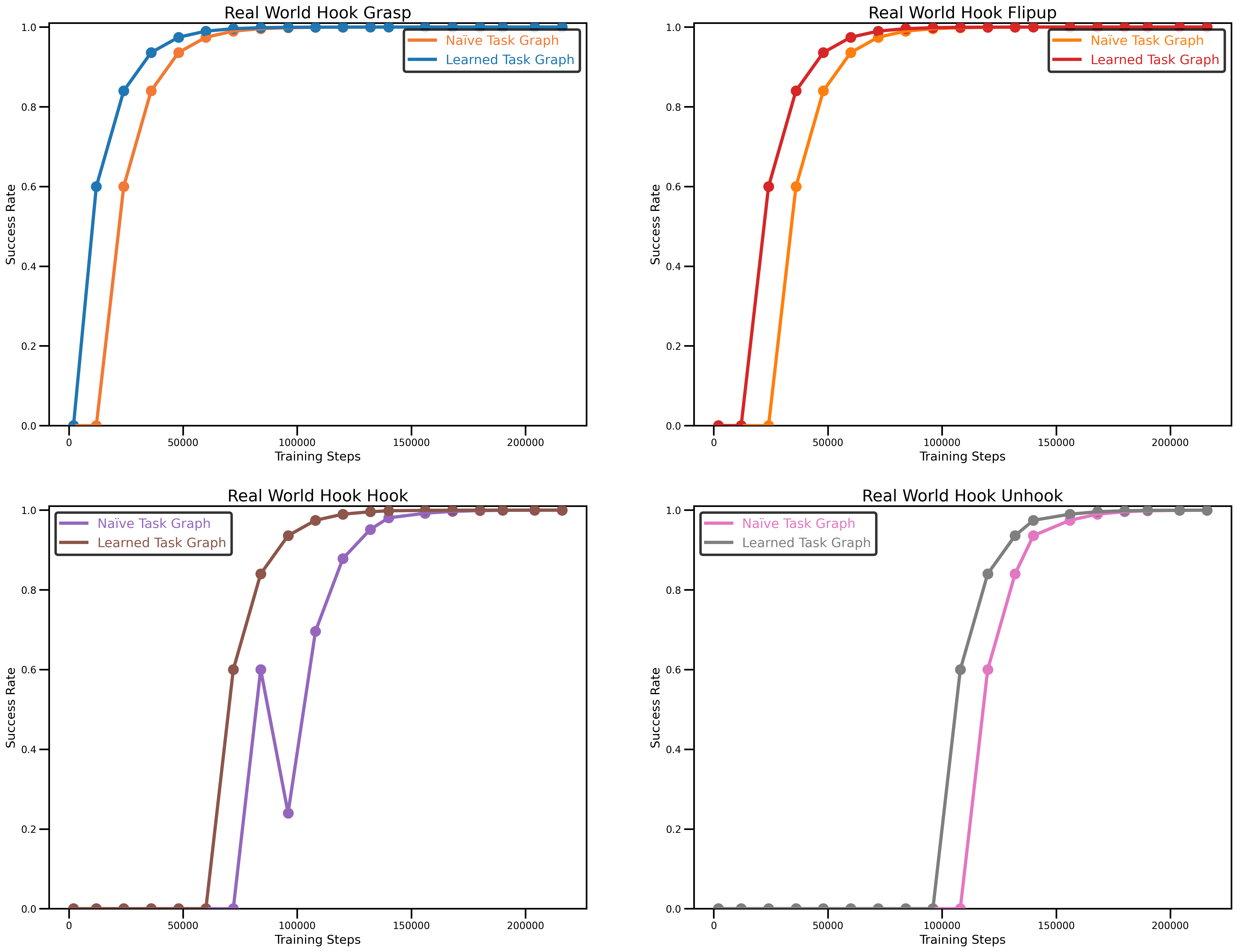}
    \vspace{-0.2cm}
    \caption{\footnotesize{Success rate of each task on our real world insertion task. We find that using a learned task graph results in faster convergence on our real world robotic task, where the robot begins to consistently perform the task around 18\% faster than the na\"{i}ve task graph.}}
\label{fig:hardware_curves}
\end{figure}

\subsection{Examples of Simulated Milestones}
\vspace{-5cm}
\begin{figure}[H]
  \centering
  \includegraphics[width=0.5\columnwidth]{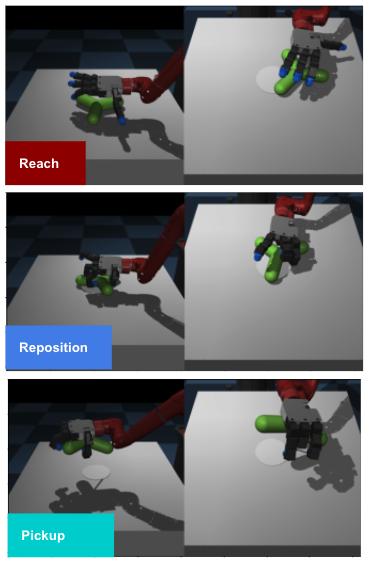}
  \caption{The experimental domain \texttt{DHandValve-v0} we study in this work. We consider a task where the simulated robot hand is required to pick up a three-pronged object. Our observations consist of images from two viewpoints (shown above), in addition to the robot's proprioceptive state. We assume no access to a ground truth reward function, nor to episodic resets. The labels in the bottom left corners were overlayed for visualization purposes.}
  \label{fig:sim-setup}
\end{figure}

\begin{figure}
  \centering
  \includegraphics[width=0.8\columnwidth]{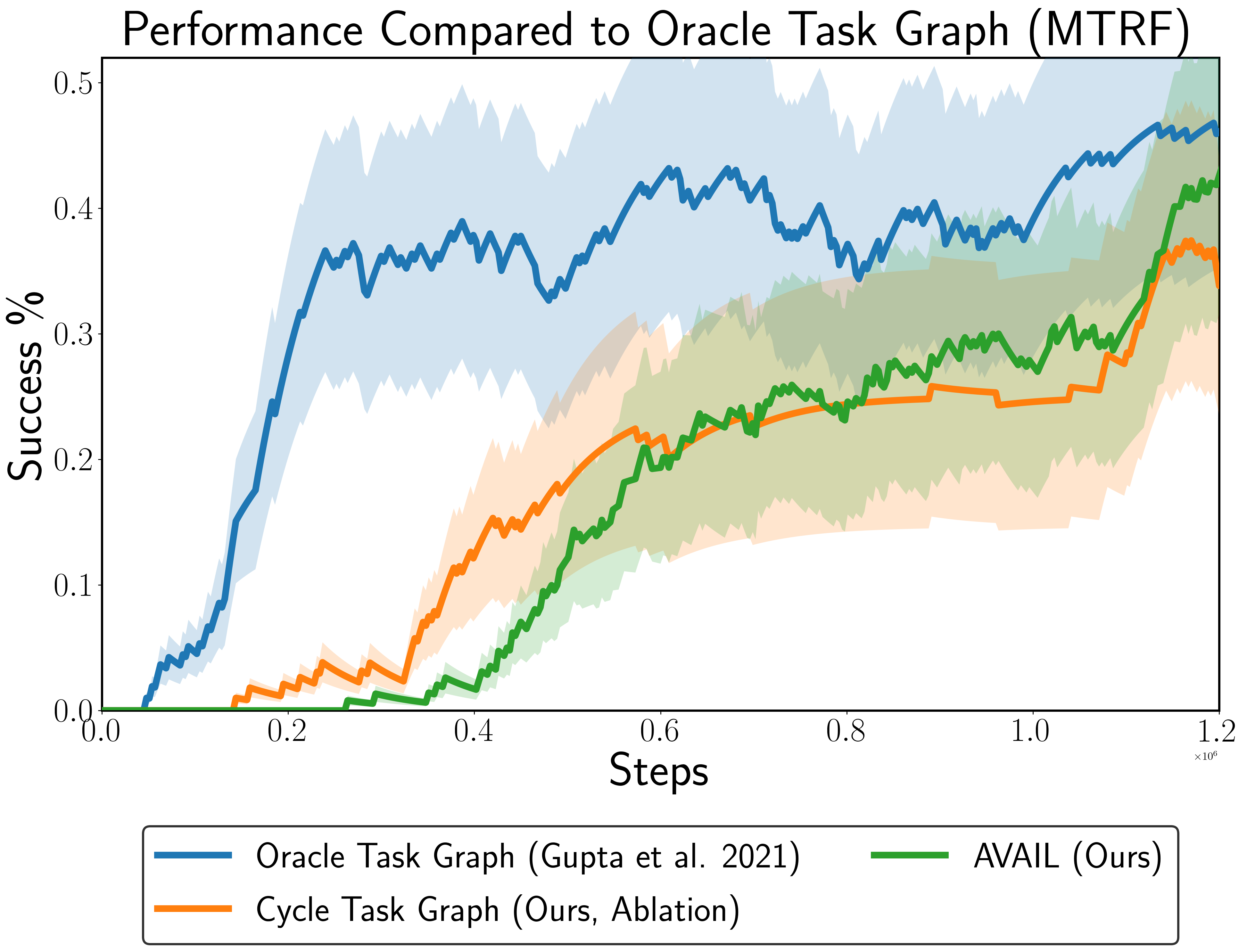}
  \vspace{-0.2cm}
  \caption{\footnotesize{A comparison of success rate on the simulated \texttt{DHandValvePickup-v0} domain compared to an oracle task graph. We find that our framework is robust to ``errors'' in the task graph compared to a hand crafted oracle. Both a learned task graph and ablation perform similarly given enough training.}}
  \label{fig:task_graph}
\end{figure}

%
%


\end{document}